\documentclass[]{interact}

\usepackage{epstopdf}
\usepackage[caption=false]{subfig}

\usepackage[longnamesfirst,sort]{natbib}
\usepackage{amssymb}

\usepackage{graphicx}%
\usepackage{algorithm2e}
\usepackage{amsmath}

\usepackage{algorithm}

\usepackage{amssymb,amsmath,amsthm}

\usepackage{multirow}%
\usepackage{amsmath,amssymb,amsfonts}%
\usepackage{amsthm}%
\usepackage{mathrsfs}%
\usepackage[title]{appendix}%
\usepackage{hyperref}
\usepackage{manyfoot}%
\usepackage{booktabs}%

\usepackage{algpseudocode}%
\usepackage{listings}%
\usepackage[utf8]{inputenc} 
\usepackage[T1]{fontenc}    
\usepackage{hyperref}       
\usepackage{url}            
\usepackage{booktabs}       
\usepackage{amsfonts}       
\usepackage{nicefrac}       
\usepackage{rotating}
\usepackage{savesym}
\usepackage{graphicx}
\usepackage{adjustbox}
\usepackage{xcolor}
\savesymbol{AND}
\usepackage{epstopdf}
\usepackage{adjustbox}
\usepackage{tabularx}
\usepackage{lscape}

\DeclareMathOperator*{\argmin}{arg\,min}

\usepackage{mathtools}
\usepackage{longtable,pdflscape,booktabs}
\DeclarePairedDelimiterX{\norm}[1]{\lVert}{\rVert}{#1}
\DeclarePairedDelimiterX{\inp}[2]{\langle}{\rangle}{#1, #2}
\usepackage{rotating}
\usepackage{longtable}
\usepackage{adjustbox}
\usepackage{booktabs} 
\usepackage{comment}

\begin{document}

\articletype{Original Research Article}

\title{A Two-Stage Interpretable Matching Framework for Causal Inference}

\author{
\name{ Sahil Shikalgar\textsuperscript{a}  and Md. Noor-E-Alam\textsuperscript{a}\thanks{CONTACT Md. Noor-E-Alam. Email: mnalam@neu.edu}}
\affil{\textsuperscript{a}Department of Mechanical and Industrial Engineering, Northeastern University, Boston, MA 02115, USA}
}

\maketitle

\begin{abstract}

Matching in causal inference from observational data aims to construct treatment and control groups with similar distributions of covariates, thereby reducing confounding and ensuring an unbiased estimation of treatment effects. This matched sample closely mimics a randomized controlled trial (RCT), thus improving the quality of causal estimates. We introduce a novel Two-stage Interpretable Matching (TIM) framework for transparent and interpretable covariate matching. In the first stage, we perform exact matching across all available covariates. For treatment and control units without an exact match in the first stage, we proceed to the second stage. Here, we iteratively refine the matching process by removing the least significant confounder in each iteration and attempting exact matching on the remaining covariates. We learn a distance metric for the dropped covariates to quantify closeness to the treatment unit(s) within the corresponding strata.  We used these high-quality matches to estimate the conditional average treatment effects (CATEs). To validate TIM, we conducted experiments on synthetic datasets with varying association structures and correlations. We assessed its performance by measuring bias in CATE estimation and evaluating multivariate overlap between treatment and control groups before and after matching. Additionally, we apply TIM to a real-world healthcare dataset from the Centers for Disease Control and Prevention (CDC) to estimate the causal effect of high cholesterol on diabetes. Our results demonstrate that TIM improves CATE estimates, increases multivariate overlap, and scales effectively to high-dimensional data, making it a robust tool for causal inference in observational data.

\end{abstract}


\section{Introduction}\label{sec1}
Observational data serve as a crucial resource for making causal inferences and informing healthcare policy decisions, especially when conducting a randomized controlled trial (RCT) is impractical due to ethical concerns or cost constraints \citep{silverman2009randomized}. However, non-randomized data in observational studies introduce challenges \citep{stroup2000meta} in estimating causal effects, as treatment assignment is often influenced by confounding factors. This lack of randomization can lead to selection bias, where certain groups receive treatment based on factors not related to their effectiveness. For example, younger patients might be more likely to receive aggressive treatments, while older patients or those with chronic conditions may be given more conservative care, making causal inference challenging. These systematic differences make it difficult to determine whether observed outcomes are due to the treatment itself or underlying factors that influenced treatment assignment.

When we understand the underlying process and control observed covariates in non-randomized data, we can reliably estimate causal effects \citep{rubin1974estimating}. Matching methods provide a powerful and interpretable solution by constructing comparable groups \citep{stuart2010matching, greenwood1945experimental, chapin1947experimental}, effectively replicating the balance achieved in randomized controlled trials (RCTs). Due to their simplicity and ease of interpretation, matching techniques have become increasingly popular among applied researchers.

Matching improves overlap in the multivariate distribution between treatment and control groups, ensuring greater overlap post-matching than pre-matching. Different methods take distinct approaches to achieving this balance. Two widely used techniques in policy evaluation are Propensity Score Matching (PSM) \citep{rosenbaum1983central} and Genetic Matching (GEN) \citep{diamond2013genetic}.
PSM seeks to improve balance by matching units with similar propensity scores. The propensity score for an individual $i$ is defined as the probability of receiving treatment given the observed covariates
(i.e., $P(T_i=1|\mathbf{X})$ where $T \in  \{0,1 \}$ is the treatment status of samples).
\cite{king2019propensity} highlight that PSM may fail to improve overall covariate balance, as it collapses the multivariate distribution into a single probability dimension. 
GEN reduces imbalance by using the smallest p-value from two-sample t-tests and Kolmogorov-Smirnov (KS) tests for each covariate. This approach evaluates balance on a variable-by-variable basis, rather than considering the joint distribution of covariates. As a result, both PSM and GEN can leave residual imbalances in the multivariate distribution, allowing substantial differences to persist across individual covariates, even among matched pairs.

When dealing with mixed data types, techniques like Coarsened Exact Matching (CEM) \citep{iacus2012causal} aim to match exactly on discrete variables while coarsening continuous variables, which is often ideal in many situations \citep{imai2008misunderstandings}. However, strictly requiring exact matches —particularly for discrete variables— can exclude a significant number of units from the analysis. This exclusion may introduce greater bias than allowing inexact matches, which retain more units and preserve a broader representation of the data \citep{rosenbaum1985constructing}. More recent methods, such as MALTS \citep{parikh2022malts}, employ weighted Hamming distance for discrete variables. However, despite these advancements, both approaches overlook the underlying distribution of discrete variables, leading to suboptimal matches.

Research on distance measures \citep{ahmad2007k} shows that the commonly used binary distance metric, defined as:
$\delta(x, y) = \mathbf{1}_{x \neq y}$ where  $\mathbf{1}_{x \neq y}$is the indicator function that equals 1 if $ x \neq y  $ and 0 if  $x = y$ are often insufficient for accurately capturing the true dissimilarity between discrete values. This metric treats all mismatches as equally distant, ignoring the actual distribution of discrete values. In contrast, effective discrete distance measures should incorporate frequency and distribution to better reflect differences between units. A purely binary approach fails to account for these differences, leading to potential mismatches and inefficiencies in the matching process.

In this work, we propose TIM (Two-stage Interpretable Matching), a method within the potential outcomes framework that accommodates binary treatments and both continuous and binary outcomes. Designed for datasets containing a mix of continuous and discrete variables, TIM offers a transparent and effective solution for causal inference by combining exact matching with iterative refinements. Unlike Coarsened Exact Matching (CEM), which often excludes a significant number of treatment units to achieve perfect balance, TIM retains more units while maintaining interpretability. Compared to Propensity Score Matching (PSM), which relies on a single scalar score, TIM's variable importance-based approach achieves superior balance across covariates.  Building on the work of \citet{ahmad2007k}, which highlights the limitations of traditional binary distance metrics that treat all mismatches as equally distant, TIM adopts a distribution-aware distance measure that considers the frequency and co-occurrence patterns of discrete variables for more accurate matching. By striking a balance between interpretability, bias reduction, and overlap improvement, TIM provides reliable Conditional Average Treatment Effect (CATE) estimates while ensuring computational scalability. Through extensive simulations and a real-world healthcare application using CDC data, we demonstrate that TIM is a robust and effective choice for causal inference in observational studies.

The paper is structured as follows: Section \ref{sec2} introduces the preliminaries, providing the necessary background for our approach. Section \ref{sec3} presents the TIM matching framework in detail. Section \ref{sec4} describes the numerical experiments conducted to evaluate our method. Section \ref{sec5} applies TIM to real-world healthcare data, demonstrating its practical performance. Section \ref{sec6} includes a discussion of our findings, and Section \ref{sec7} concludes the paper with key takeaways and future directions.

\section{Preliminaries}\label{sec2}

\subsection{Setting}\label{sec2.1}
In this paper, we adopt the potential outcomes framework \citep{holland1986stat} and explore matching methods developed within this framework to identify treatment effects from observational data.

Consider a sample of \( n < N \) units drawn from a population of size \( N \). Let \( T_i \) be an indicator variable for unit \( i \), where \( T_i = 1 \) if the unit belongs to the ``treated'' group and \( T_i = 0 \) if it belongs to the ``control'' group. The observed outcome variable is defined as  
$Y_i = T_i Y_i(1) + (1 - T_i) Y_i(0)$,
where \( Y_i(1) \) represents the potential outcome for unit \( i \) if it receives treatment, and \( Y_i(0) \) otherwise. 

Matching methods aim to recover the underlying randomization by controlling for pretreatment covariates. To formalize this, let \( X_i = (X_{i1}, X_{i2}, \dots, X_{ik}) \) represent the \( k \)-dimensional covariate vector for unit \( i \), which includes both continuous and discrete variables. Specifically, we define:  
\(X_{ik} = \{ X_{ik_c}, X_{ik_d} \}\)
where \( |X_{k_c}| \) represents the subset of \( p_c \) continuous covariates, and \( |X_{k_d}| \) represents the subset of \( p_d \) discrete covariates, with \( p_c + p_d = k \).  
The full covariate matrix for the \( n \) sampled units, which may be drawn from a larger population of size \( N \), is then expressed as:  
\(
\mathbf{X} = [X_{ij}, \, i = 1, \dots, n, \, j = 1, \dots, k]
\)

Let \( \mathcal{T} = \{ i: T_i = 1 \} \) be the set of indices for the treated units, with \( n_T = \mathcal{|T|} \) representing the number of treated units. Similarly, let \( \mathcal{C} = \{ i: T_i = 0 \} \) be the set of indices for the control units, with \( n_C = |\mathcal{C}| \) representing their count, such that \( n_T + n_C = n \).

The total number of strata, denoted as $ \mathcal{S}$, is determined by the number of unique covariate combinations among the treatment units $\mathcal{T}$ in the dataset. Formally, let each stratum \( s \) correspond to a unique combination of covariates associated with a treatment unit. Each stratum \( s \) contains all control units whose covariate values match the treatment unit defining that stratum, ensuring that:  
\(
s \in \{1, \dots, \mathcal{S}\}, \text{with each } s \text{ containing at least one treatment and one control unit}.
\)

We define \( m_T^s \) and \( m_C^s \) as the number of treated and control units matched within each stratum \( s \), considering both continuous and discrete covariates during the matching process. Let \( M_T^s \subset T \) and \( M_C^s \subset C \) represent the sets of indices corresponding to the matched treated and control units in stratum \( s \).  
The total number of matched treated and control units across all strata satisfies the following constraint:  
\(
\sum_{s=1}^{\mathcal{S}} m_T^s \leq n_T, \quad \sum_{s=1}^{\mathcal{S}} m_C^s \leq n_C,
\)
where \( \mathcal{S} \) denotes the total number of strata. These inequalities hold because not all treated and control units necessarily have suitable matches, meaning some units may remain unmatched.

 Throughout the remaining manuscript, we make the following standard assumptions: 
\begin{enumerate}
    \item Stable Unit Treatment Value \citep{rubin2005causal}
    \item Strong ignorability (\citep{rosenbaum1983central}: \( Y(1) \perp Y(0) \mid T \mid X \)
\end{enumerate}

Based on our assumptions, the individual causal effect is \( D = Y_i(1) - Y_i(0) \), and the Conditional Average Treatment Effect (CATE) is:
$\tau(x) = \mathbb{E}\left\{ Y(1) - Y(0) \mid\textbf{X}  \right\} = \mathbb{E}(D \mid \textbf{X} )$.
\subsection{Variable Importance}\label{sec2.2}
The set of covariates ($\textbf{X}$) can be further categorized based on their association with treatment and/or outcome. In parametric approaches to causal effect estimation, two key statistical models are considered: the outcome model and the treatment allocation mechanism, both expressed as functions of pre-treatment variables \citep{ertefaie2018penalization}.

To ensure the assumption of ignorable treatment assignment holds, the matching procedure must include all variables that influence both treatment assignment and the outcome \citep{glazerman2003nonexperimental, rubin1996matching}. These variables, known as confounders, must be accounted for in the matching process. By doing so, we ensure that any differences in outcomes between treated and control units can be attributed to the treatment effect rather than confounding factors. Omitting an important confounder can significantly increase bias, so researchers should err on the side of inclusivity when selecting variables associated with treatment assignment and/or outcome.

Most existing methods prioritize covariate importance through outcome modeling \citep{parikh2022malts, wang2021flame}. However, this approach has a major drawback: it often overlooks confounders that are strongly related to treatment but weakly related to outcome, despite their relevance. These confounders remain non-ignorable and must be accounted for.

To address these challenges, we determine confounder importance ($\boldsymbol\theta^* $) directly from the data using two linear models—one for treatment ($\hat{\boldsymbol\alpha}$) and one for outcome ($\hat{\boldsymbol\beta}$). By averaging their results, we obtain a more balanced measure of confounder significance, mitigating biases introduced by a single modeling approach as shown in Algorithm \ref{algo:1}.

\begin{algorithm}
\caption{Algorithm to estimate $\theta^*$ using Regression}\label{algo:1}  
 
\textbf{Input:} Dataset $\textbf{X}, \textbf{Y} =\sum_{i=1}^nY_i ,T_i$ \\

 \textbf{Step 1:}   \Comment{Run linear regression Outcome Model to get coefficients}

 $\hat{\boldsymbol{\beta}}= \argmin_{\boldsymbol\beta} l_n(\boldsymbol{\beta}; \textbf{Y}, \mathbf{X}) \textrm{ (MLE of $\beta$) }$\\
     
 \textbf{Step 2:}     \Comment{Run logistic regression for binary Treatment Model to get coefficients}
    
$        \hat{\boldsymbol{\alpha}} =   {\argmin}_{\boldsymbol\alpha} \biggl \{ \sum_{i=1}^n  ( -a_iX_i^{T_i}\alpha) +
   log(1+e^{X_i^{T_i}\alpha})  ) \biggr \} $\\
\textbf{Step 4:}    Normalize $\hat{\boldsymbol{\beta}},  \hat{\boldsymbol{\alpha}} $\\
\textbf{Step 5:}         \Comment{Average element-wise absolute values of coefficients}

 $\boldsymbol\theta^* \gets \{\hat{\boldsymbol{\beta}} + \hat{\boldsymbol{\alpha}}\}/2 $ \\
{\textbf{Result:} Vector of Confounder Importances $\boldsymbol\theta^*$}\\

\end{algorithm}
\subsubsection{Alternative Methods}  

While we primarily use simple linear regression and logistic regression for ease of understanding—particularly in the context of public health research—more advanced machine learning models can be employed for finding $\boldsymbol\theta^* $.  

If researchers wish to explore more complex methods for assessing $\boldsymbol\theta^* $, tree-based approaches offer useful alternatives. For instance, when using decision trees, variable importance can be determined via Gini importance, feature permutation importance, or similar metrics. More broadly, for any model class, subtractive model reliance can be applied to measure how the loss function changes when a given covariate is perturbed \citep{fisher2019all}.

\subsection{Unified Distance Measure}\label{sec2.3}

In this section, we introduce a mixed distance measure that integrates both continuous and discrete variables. The covariate set \( \mathbf{X} \) can be divided into two components: continuous variables (\( X_{k_c} \)) and discrete variables (\( X_{k_d} \)).  
Our goal is to identify the closest control units based on the covariate distance to the treatment unit(s) within each stratum \( s \) for effective matching.

Measuring the distance between continuous variables is straightforward (e.g., the difference between \( 5 \, \text{m} \) and \( 3 \, \text{m} \) is \( 2 \, \text{m} \)). However, defining distances for categorical variables is more complex. For example, how do we quantify the distance between categories such as \textit{Blue}, \textit{Green}, and \textit{Red}? A common approach for measuring distance between categorical variables is the binary distance metric, \( \delta(p, q) \), which assigns a value of 1 if \( p \neq q \) and 0 if \( p = q \). However, research by \cite{stanfill1986toward} suggests that in supervised learning, while \( \delta(p, q) = 0 \) when \( p = q \), it is not necessarily true that \( \delta(p, q) = 1 \) when \( p \neq q \). Instead, \( \delta(p, q) \) can vary depending on the relative frequencies of attribute pairs in relation to other covariates. Although this approach was originally designed for clustering, it can be adapted for causal inference by treating treatment units as cluster centers.

In the following, we draw inspiration from \cite{ahmad2007k} and modify their proposed mixed-data cost function for clustering to be applied in the matching process. For more information, please refer to the cited article. Equation \ref{eq:component_function} illustrates how to learn a mixed distance metric between a treatment unit and a control unit within the same stratum.

\begin{equation}
    \Delta(\mathcal{T}_i, \mathcal{C}_j) = \sum_{k_c=1}^{p_{c}} \left( X_{ik_c} - X_{jk_c}\right)^2 
    + \sum_{k_d=1}^{p_d} \Omega(X_{ik_d},X_{jk_d})^2.
    \label{eq:component_function}
\end{equation}

In Equation~(\ref{eq:component_function}), the term  
\(
\sum_{k_c=1}^{p_{c}} \left( X_{ik_c} - X_{jk_c} \right)^2
\) 
represents the distance between the control unit \( \mathcal{C}_j \) and its corresponding treatment unit \( \mathcal{T}_i \) based solely on numeric attributes, which is computed using the Euclidean distance.  
The term  
\(
\sum_{k_d=1}^{p_d} \Omega(X_{ik_d}, X_{jk_d})^2
\) 
measures the distance between the control unit \( \mathcal{C}_j \) and its corresponding treatment unit \( \mathcal{T}_i \) based on discrete attributes. In the following section, we formally define how \( \Omega(x, y) \) is computed for discrete variables.

Let \( X_{k_{d1}} \) denote a discrete attribute with values \( x \) and \( y \). To determine the distance between \( x \) and \( y \), we consider their overall distribution in the dataset along with their co-occurrence with values of other attributes. Suppose \( X_{k_{d2}} \) denotes another discrete attribute, and let \( w \) be a subset of values of \( X_{k_{d2}} \). Using set-theoretic notation, \( \bar{w} \) denotes the complementary set of values occurring for attribute \( X_{k_{d2}} \).

 \textbf{Definition 1 }
For a dataset with \( k \) attributes, including both discrete and discretized numeric attributes, the overall distance between two distinct values \( x \) and \( y \) of any discrete attribute \( X_{k_{d1}} \) is given by:  
\begin{equation}
    \Omega(x, y) = \frac{1}{k - 1} \sum_{\substack{j=1 \dots k \\ i \neq j}}^{m} \delta_{ij}(x, y).
    \label{eq:distance3}
\end{equation}

where

\begin{equation}
    \delta_{ij}(x, y) = P_i(\omega\mid x) + P_i(\bar{\omega} \mid y) - 1.0.
    \label{eq:distance2a}
\end{equation}

Where \( \omega \) is the subset \( w \) of values of \( X_{k_{d1}} \) that maximizes the quantity \( P_i(x \mid x) + P_i(\bar{x} \mid y) \). Since both \( P_i(x \mid x) \) and \( P_i(\bar{x} \mid y) \) lie between 0 and 1, to ensure that \( \delta_i(x,y) \) is restricted to the range [0,1] we subtract 1.0 from Equation \ref{eq:distance2a}.

Equation~(\ref{eq:distance2a}) expresses the distance between values \( x \) and \( y \) of \( X_{k_{d1}} \) as a function of their co-occurrence probabilities with a set of values of another discrete attribute \( X_{k_{d2}} \). By selecting the subset \( x \) that maximizes this value, it aims to capture the maximum contribution that \( x \) and \( y \) can make toward the distance if \( X_{k_{d1}} \) is the only other attribute. 

When additional discrete attributes are present, similar distance measures for the pair \( x \) and \( y \) can be computed for each attribute individually, as shown in Equation~(\ref{eq:distance3}).  
 The absolute distance between \( x \) and \( y \) is then obtained as the average of these values. The distance between \( x \) and \( y \) with respect to a numeric attribute is computed by first discretizing the attribute. However, this discretized form is only used for computing discrete covariate distances.

\subsection{Imbalance Measure}\label{sec2.4}

The purpose of measuring imbalance is to quantify the difference between the multivariate empirical distributions of the pretreatment covariates for the treated group, \( \mathcal{T} \), and the control group, \( {\mathcal{C}} \). An approach introduced by \citet{iacus2011multivariate} evaluates these multivariate differences using the \( L_1 \) distance.

Let \( H(X_1) \) denote the set of distinct bins for the variable \( X_1 \), which partitions its distribution into discrete intervals. The multidimensional histogram is then constructed as the Cartesian product of these partitions across all \( k \) variables:  
\(
H(X) = H(X_1) \times H(X_2) \times \dots \times H(X_k).
\)
We define \( f \) and \( g \) as the relative empirical frequency distributions for the treated and control groups, respectively. The relative frequency of observations within a given histogram cell \( (\ell_1, \ell_2, \dots, \ell_k) \) is represented as \( f_{\ell_1\ell_2\dots\ell_k} \) for the treated group and \( g_{\ell_1\ell_2\dots\ell_k} \) for the control group.

\textbf{Definition 2} \citep{iacus2011multivariate}:  

The multivariate imbalance measure is given by  

\begin{equation}
L_1(f, g) = \frac{1}{2} \sum_{\ell_1\dots\ell_k \in H(X)} \left| f_{\ell_1\dots\ell_k} - g_{\ell_1\dots\ell_k} \right|.
\end{equation}

Let \( f^m \) and \( g^m \) represent the empirical frequency distributions for the matched treated and control groups, respectively, using the same discretization as the unmatched distributions \( f \) and \( g \). An effective matching method should ensure that  
\(
L_1(f^m, g^m) < L_1(f, g).
\)  
The \( L_1 \) value provides an intuitive measure of imbalance. If the two distributions are completely disjoint—given the granularity of the histogram—then \( L_1 = 1 \). Conversely, if they exhibit perfect overlap, then \( L_1 = 0 \). For cases where the distributions partially overlap, the \( L_1 \) value falls within the range \( (0,1) \).

\section{Two-stage Interpretable Matching (TIM) framework} \label{sec3}
Working with high-dimensional data makes exact matching impractical. This is due to the sparsity of data in high-dimensional spaces, a challenge known as the curse of dimensionality. As \cite{rosenbaum1985bias} pointed out, insisting on exact matches can leave many individuals unmatched, potentially introducing greater bias than allowing inexact matches while retaining more data for analysis. To address this issue, we propose TIM, a hybrid approach that combines exact matching with distance-based matching for mixed data.

TIM employs a structured matching approach by first attempting to match treatment and control units exactly across all variables. This ensures that all possible exact matches are identified before introducing any flexibility. If exact matching is not feasible for some treatment units, the process is iteratively refined by removing the least important variable each iteration. This creates matched strata based on \(k-u \) variables, where \( k \) is the total number of covariates and \( u\) represents the number of iterations. With each iteration, \( u \) increases by 1, and the process continues until all treatment units have found a suitable match.  

Within each matched stratum, we compute the mixed distance defined in Equation~(\ref{eq:component_function}) using the \( u \) variables where exact matches were not found. This additional refinement ensures the best possible pairing between treatment and control units within each stratum.
Our methodology consists of three key phases: Preprocessing, Matching, and Refining Matched Strata, each designed to balance matching quality and sample retention effectively.

\begin{figure}[h]
    \centering
    \includegraphics[width=0.8\textwidth]{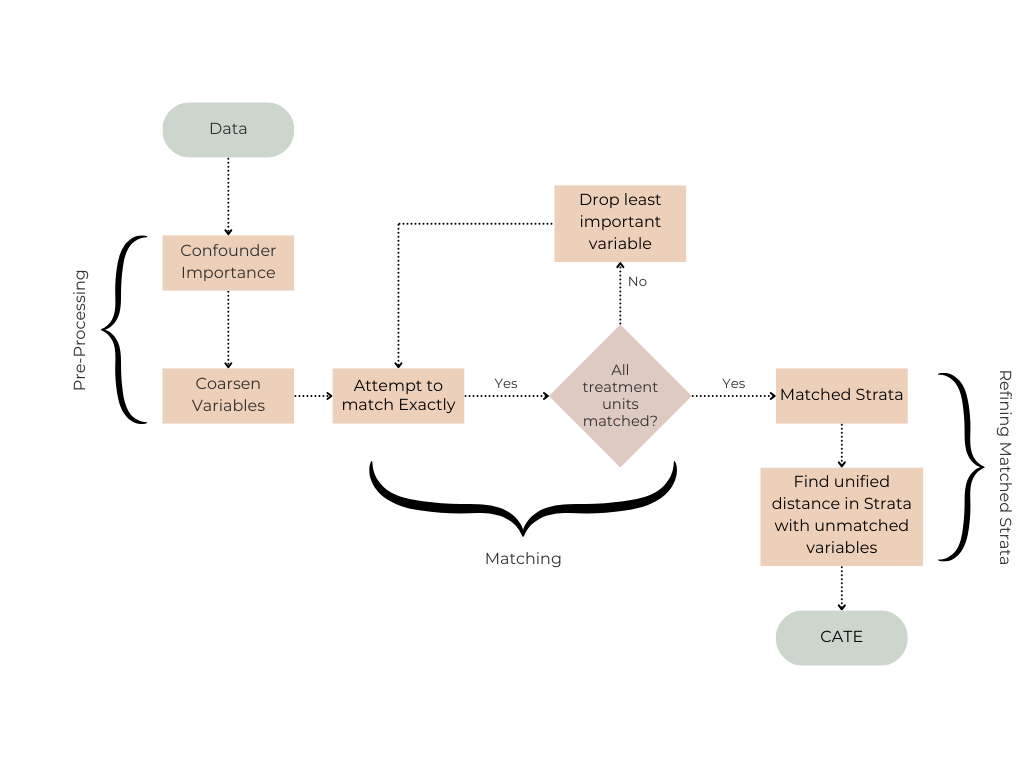}
    \caption{Flowchart of TIM}
    \label{fig:TIMv2}
\end{figure}


In the Preprocessing phase, as illustrated in Figure~\ref{fig:TIMv2}, we begin by evaluating the importance of confounders using Algorithm~\ref{algo:1}. This step identifies the least important variables so that they can be removed first during the matching process. Next, we coarsen continuous variables using techniques from Coarsened Exact Matching (CEM) \citep{iacus2012causal}, reducing the granularity of the data.

The Matching phase follows a structured, iterative process designed to maximize the number of exact matches while minimizing bias, as outlined in Algorithm~\ref{algo:2}. Initially, in iteration (0), we attempt to match treatment and control units exactly across all \( k \) covariates.  
We construct \( \mathcal{G} \), representing the initial strata, and then refine \( \mathcal{G} \) to obtain \( s \), which contains at least one treatment unit (\( m^s_T \)) and one control unit (\( m^s_C \)).  
These refined strata are then appended to \( \mathcal{S} \).  
Next, we remove the matched treatment units (\( M^s_T \)) and control units (\( M^s_C \)) from \( X_{\text{remaining}} \), ensuring that they are not reconsidered in subsequent matching iterations.
If exact matches cannot be found for some treatment units, we initiate an iterative refinement process. 
In each iteration, we remove the least significant variable, determined by the smallest value of \( \boldsymbol\theta^* \), 
and attempt matching using the remaining \( k - u \) variables, where \( u \) represents the number of iterations or variables removed. 
We then follow the same procedure as before to identify \( s \) and remove matched units from \( X_{\text{remaining}} \), 
ensuring they are not reconsidered in subsequent iterations.
This process continues until all treatment units are matched, as outlined in Step 2 of Algorithm~\ref{algo:2}. 
After this phase, we obtain the set of Matched Strata \( \mathcal{S} \), 
which consists of strata where exact matches were achieved across all \( k \) variables, 
as well as strata where matches were made based on fewer than \( k \) variables.
 By progressively relaxing the matching constraints, we ensure that as many units as possible are retained in the analysis without compromising interpretability.

\begin{algorithm}[H]
\caption{Exact Matching with Variable Importance}
\label{algo:exact_matching}
\KwIn{$\textbf{X}$, $\textbf{T} = \sum_{i=1}^n T_i$, $\boldsymbol{\theta}^*$}
\KwOut{Matched strata ${\mathcal{S}}$, unmatched treated units $M_{T_u}$, remaining unmatched control units $M_{C_u}$}

\textbf{Step 1: Sort Covariates by Importance} \\
$k^* \gets$ indices of covariates sorted by $\boldsymbol{\theta}^*$ in descending order \\
$X_{\text{remaining}} \gets \textbf{X}$

\vspace{0.5em}
\textbf{Step 2: Iterative Matching Process} \\
\For{$k$ from $k^*$ down to $1$}{
    $\mathcal{G} \gets$ group $X_{\text{remaining}}$ by $k$ covariates \\
    
    \ForEach{$s \in \mathcal{G}$}{
        \If{$m_T^s > 1$ \textbf{and} $m_C^s > 1$}{
            Append $s$ to $\mathcal{S}$ \\
            Remove matched indices $M_T^s$, $M_C^s$ from $X_{\text{remaining}}$
        }
    }

    Extract remaining treated units: $M_{T_u} \gets M_{T_u} - \bigcup_s M_T^s$ \\
    \If{$M_{T_u}$ is empty}{
        \textbf{break}
    }
}

\vspace{0.5em}
\textbf{Step 3: Return Results} \\
\Return ${\mathcal{S}}, M_{T_u}, M_{C_u}$
\end{algorithm}

In the Refining Matched Strata phase, we address the limitations of exact matching by introducing a distance-based refinement step. For treatment units that could not be matched exactly on all covariates, we learn a mixed distance metric to assess the closeness between treatment and control units within each matched stratum $s$. This refinement allows us to incorporate previously unmatched variables $u$ in a principled manner, ensuring that the final matches are as close as possible.

As shown in Algorithm~\ref{algo:3}, we compute the continuous distance \( \Delta^s_{c} \) and discrete distance \( \Delta^s_{d}\) for the unmatched variables \( u \) within stratum \( s \). The grand total distance \( \Delta^s \) in stratum \( s \) is then calculated over all unmatched variables, denoted as \( X^s_{k_{cu}} \) for continuous and \( X^s_{k_{du}} \) for discrete variables.  

Once the grand total distance for each control unit relative to the treatment unit in stratum where \( u \geq 1 \) is obtained, we compute \( \textbf{I}^s \), an inverse score that assigns a value of 1 to the closest control unit and 0 to the most distant control unit(s) within the stratum. For strata where exact matches are found across all covariates, we set \( \textbf{I}^s = 1 \).
The inverse score \( \textbf{I}^s \) is then used to weight the estimation of the Conditional Average Treatment Effect (CATE), ensuring that control units closer to the treatment unit receive higher weights in the estimation.

\begin{algorithm}[H]
\caption{Unified Distance Calculation for Matched Strata}
\label{algo:3}
\KwIn{${\mathcal{S}}$, $\textbf{T}$, $\textbf{Y}$, $X_{k_c}$, $X_{k_d}$}
\KwOut{Updated matched datasets with calculated unified distances ${\mathcal{S}}^*$}

\ForEach{$s \in {\mathcal{S}}$}{
    $u \gets$ unmatched covariates from each stratum \tcp*{Set of unmatched covariates}
    
    $X^s_{k_{cu}} \gets X_{k_c} \cap$ continuous unmatched covariates \\
    $X^s_{k_{du}} \gets X_{k_d} \cap$ discrete unmatched covariates \\
    
    \If{$u \neq \emptyset$}{
        
        \tcc{Compute Continuous Distances}
        \ForEach{$x^s_{k_{cu}} \in X^s_{k_{cu}}$}{
            $\delta^s_{c}(\mathcal{T}_{i}, \mathcal{C}_{j}) = \sum_{j=1}^{m^s_{C}} \left( x_{ik_{uc}} - x_{jk_{uc}}\right)^2$
        }
        $\Delta^s_{c} \gets \delta^s_{c}$

        \tcc{Compute Discrete Distances}
        \ForEach{$x^s_{k_{du}} \in X^s_{k_{du}}$}{
           $\delta^s_{d}(\mathcal{T}_{i}, \mathcal{C}_{j}) = \sum_{j=1}^{m^s_{C}} \Omega\left( x_{ik_{ud}}, x_{jk_{uc}}\right)^2$
        }
        $\Delta^s_{d} \gets \delta^s_{d}$

        \tcc{Compute Grand Total Distance}
        $\Delta^s = \Delta^s_{c} +  \Delta^s_{d}$

        \tcc{Apply Inverse Min-Max Normalization}
        $\mathbf{I}^s = 1 - \frac{\Delta^s - \Delta^s_{\min}}{\Delta^s_{\max} - \Delta^s_{\min}}$ \\
        \tcp{Assign 1 to missing values for control units}

        \tcc{Update $\mathcal{S}$ with computed distances}
        $s^* \gets$ $s$ with $\Delta^s$ and $\mathbf{I}^s$ \\
        $\mathcal{S}^* \gets \mathcal{S}^* \cup \{s^*\}$
    }
}
\Return $\mathcal{S}^*$
\end{algorithm}

By following this structured approach, TIM effectively balances the trade-off between exact matching quality and sample retention. This method ensures that we maximize the number of matched individuals while mitigating bias, providing a robust and interpretable solution for matching problems. Ultimately, TIM enables the use of nearly all available treatment units in the analysis, making it a practical approach for causal inference in complex datasets.

\subsection{Estimating CATE from Strata}  

In this section, we describe the computation of the Conditional Average Treatment Effect (CATE) within each stratum.  
For a given stratum \( s^* \), if it belongs to the set where \( u = 0 \) (i.e., exact matches on all covariates are found), we compute the individual treatment effect (ITE) in the standard manner:  
\(
D = Y_i(1) - Y_i(0)
\) 
we then proceed to calculate the CATE.  
However, for strata where \( u \geq 1 \) (i.e., exact matches are not found for all covariates), we must account for the distance between control and treatment units within the stratum. To achieve this, we adjust the individual treatment effect by incorporating the inverse score \( \textbf{I}^s \), yielding:  
\(
D = I^s_i (Y_i(1) - Y_i(0))
\)  
This ensures that control units closer to the treatment units within stratum \( s \) receive higher weight.  
Next, we compute the CATE for each stratum by taking the weighted average of the ITE. Finally, to obtain the overall CATE, we average the CATE values across all strata $s^*$.

\section{Numerical Experiments} \label{sec4}
In this section, we evaluate the performance of TIM against other matching methods. Our comparison focuses on three key aspects: bias, the computational time required to estimate CATE, and the multivariate overlap of variables ($L_1$) before and after matching.

\subsection{Simulation Design}\label{sec4.1}

We designed the simulation study based on the data generation process described by \citet{shortreed2017outcome}. Our study explores three scenarios, where continuous covariates are drawn from a multivariate Gaussian distribution, and discrete covariates follow a Bernoulli distribution. The treatment variable is binary, and the outcome is continuous.  
The three scenarios differ in their association structures. For all scenarios, we set the sample size to \( n = 500 \). Each scenario is further divided into two sub-scenarios (A and B), which vary in the correlation among covariates. Specifically, we consider two cases: one where covariates are independent (\(\rho = 0\), indicating no correlation) and another where covariates are strongly correlated (\(\rho = 0.5\), representing high correlation).

The data-generating process is as follows $X_{ik} = \{X_{i{k_c}}, X_{i{k_d}} \}$ where $X{i{k_c}}$ are the continuous covariates and $X_{i{k_d}}$ are the discrete covariates:

\begin{itemize}

    \item Continuous Variables: Covariates: $X_{c_1}, X_{c_2}, \cdots,X_{c_5} \stackrel{iid}{\sim} N(0,1)$
        
    \item Categorical Variables: Binary categorical variables $X_{d_1}, X_{d_2}, X_{d_3}$ are generated using a two-step process:
        \begin{enumerate}
            \item First, for each observation $i$, a probability $p_i$ is drawn from a uniform distribution:  
            $p_i \sim \text{Uniform}(0.3, 0.7)$.
            \item Then, the categorical variable is sampled from a Bernoulli distribution with probability $p_i$:  
            $X_{d_1}, X_{d_2}, X_{d_3} \stackrel{iid}{\sim} \text{bernoulli}(p_i)$.
            
        \end{enumerate}

\end{itemize}
\textbf{Scenario 1:} Includes strong confounders.
\begin{itemize}
    \item Treatment Assignment: $T \sim Bernoulli( p = Expit(0.8X_{c_1} + \cdots + 0.8X_{c_5} +0.8X_{d_1}+ 0.8X_{d_2} + 0.8X_{d_3}))$ where, $Expit (x) = \frac{e^x}{1+e^x} $
    \item Outcome Model: The outcome variable $Y$ is generated as
    $y = TE.T + 0.8X_{c_1} + \cdots + 0.8X_{c_5} +0.8X_{d_1}+ 0.8X_{d_2} + 0.8X_{d_3} + \epsilon$,
    where $\epsilon \sim \mathcal{N}(0, \sigma_x)$ represents random noise and $TE = 1$ is the treatment effect and $T=0$ when unit does not receive treatment, $T=1$ when it receives treatment.
    \item Correlation:  $\rho = 0$ (Scenario 1A),$\rho = 0.5$ (Scenario 2B) 
\end{itemize}

\textbf{Scenario 2:}  Includes confounders with varying strengths—strong, medium, and weak.
\begin{itemize}
    \item Treatment Assignment: $T \sim Bernoulli( p = Expit(0.8X_{c_1} + 0.8X_{c_2}+0.5X_{c_3}+0.5X_{c_4} + 0.2X_{c_5} +0.8X_{d_1}+ 0.5X_{d_2} + 0.2X_{d_3})$ where, $Expit (x) = \frac{e^x}{1+e^x} $
    \item Outcome Model: The outcome variable $Y$ is generated as
    $y = TE.T + 0.8X_{c_1} + 0.8X_{c_2}+0.5X_{c_3}+0.5X_{c_4} + 0.2X_{c_5} +0.8X_{d_1}+ 0.5X_{d_2} + 0.2X_{d_3} + \epsilon$,
    where $\epsilon \sim \mathcal{N}(0, \sigma_x)$ represents random noise and $TE = 1$ is the treatment effect and $T=0$ when unit does not receive treatment, $T=1$ when it receives treatment.
    \item Correlation:  $\rho = 0$ (Scenario 2A),$\rho = 0.5$ (Scenario 2B) 
\end{itemize}

\textbf{Scenario 3:}  Includes strong confounders, medium confounders, and confounders that are strongly associated with the treatment.
\begin{itemize}
    \item Treatment Assignment: $T \sim Bernoulli( p = Expit(0.8X_{c_1} + 0.8X_{c_2}+0.5X_{c_3}+0.5X_{c_4} + 0.8X_{c_5} +0.8X_{d_1}+ 0.5X_{d_2} + 0.8X_{d_3})$ where, $Expit (x) = \frac{e^x}{1+e^x} $
    \item Outcome Model: The outcome variable $Y$ is generated as
    $y = TE.T +0.8X_{c_1} + 0.8X_{c_2}+0.5X_{c_3}+0.5X_{c_4} + 0.2X_{c_5} +0.8X_{d_1}+ 0.5X_{d_2} + 0.2X_{d_3} + \epsilon$,
    where $\epsilon \sim \mathcal{N}(0, \sigma_x)$ represents random noise and $TE = 1$ is the treatment effect and $T=0$ when unit does not receive treatment, $T=1$ when it receives treatment.
    \item Correlation:  $\rho = 0$ (Scenario 3A),$\rho = 0.5$ (Scenario 3B) 
\end{itemize}

\subsection{Numerical Experiment Results}
We evaluate TIM using state-of-the-art methods by calculating bias, multivariate overlap, and the time taken to run the algorithms. In this experiment, our goal is to minimize bias while maximizing the overlap between the pre-match (\( L_1 \)) and post-match (\( L_1^m \)) distributions. As highlighted by \cite{iacus2009cem}, all matching methods must ensure that \( L_1^m < L_1 \) after matching, 
signifying an improvement in balance. This ensures that the matching process effectively attempts to mimic the conditions of 
a randomized controlled trial.
 We generate 1000 datasets for each scenario and apply various matching techniques to estimate the Conditional Average Treatment Effect (CATE). In all experiments, we calculate bias as the difference between the estimated CATE and the true treatment effects across 1000 iterations. Additionally, we compute the fraction of treatment units matched (\( T_{f} \)) by each matching technique to measure the proportion of successfully matched units.
 The experiments were conducted on a Dell Vostro workstation equipped with an Intel Core i5-11400H CPU running at 2.70 GHz and 16 GB of RAM.

\begin{figure}

\subfloat[]{%
\resizebox*{8 cm}{!}{\includegraphics{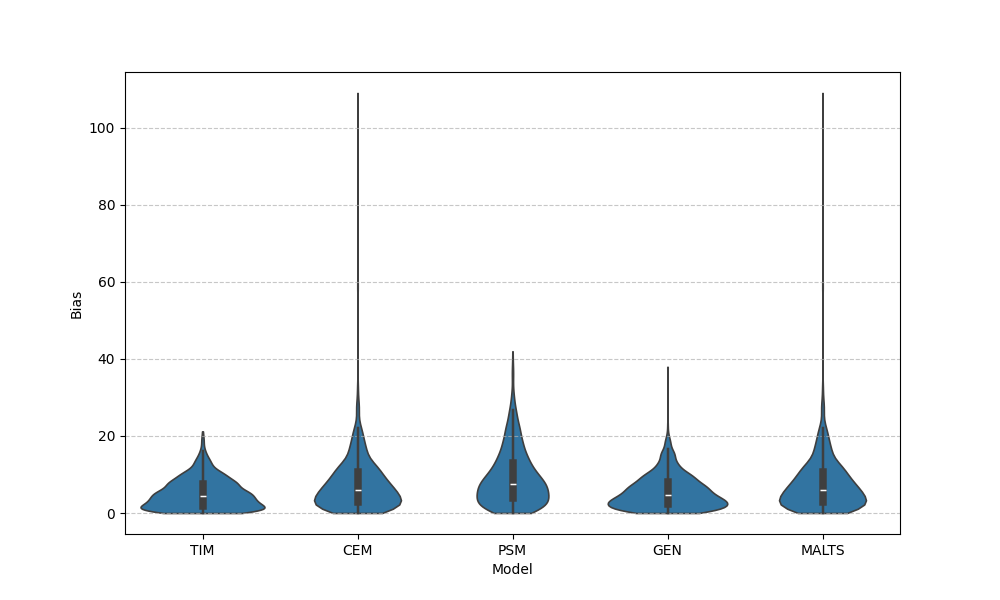}}}
\subfloat[]{%
\resizebox*{8 cm}{!}{\includegraphics{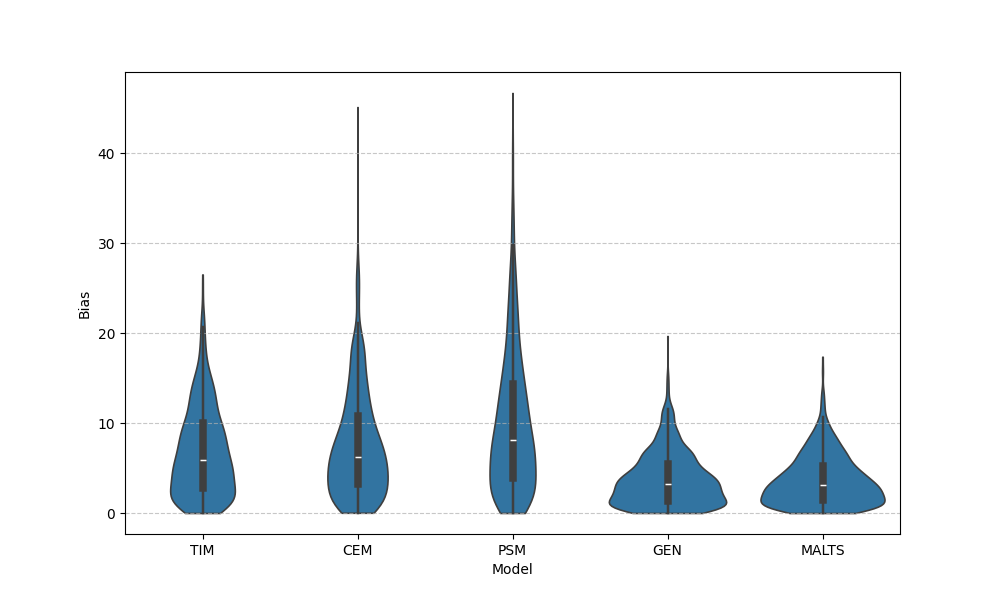}}}\hspace{5pt}
\subfloat[]{%
\resizebox*{8 cm}{!}{\includegraphics{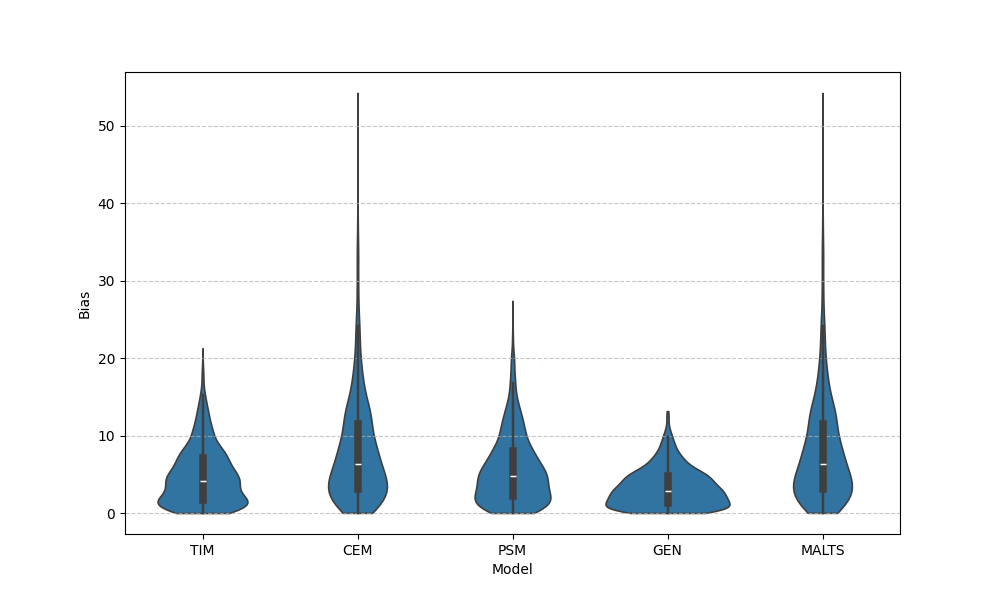}}} 
\subfloat[]{%
\resizebox*{8 cm}{!}{\includegraphics{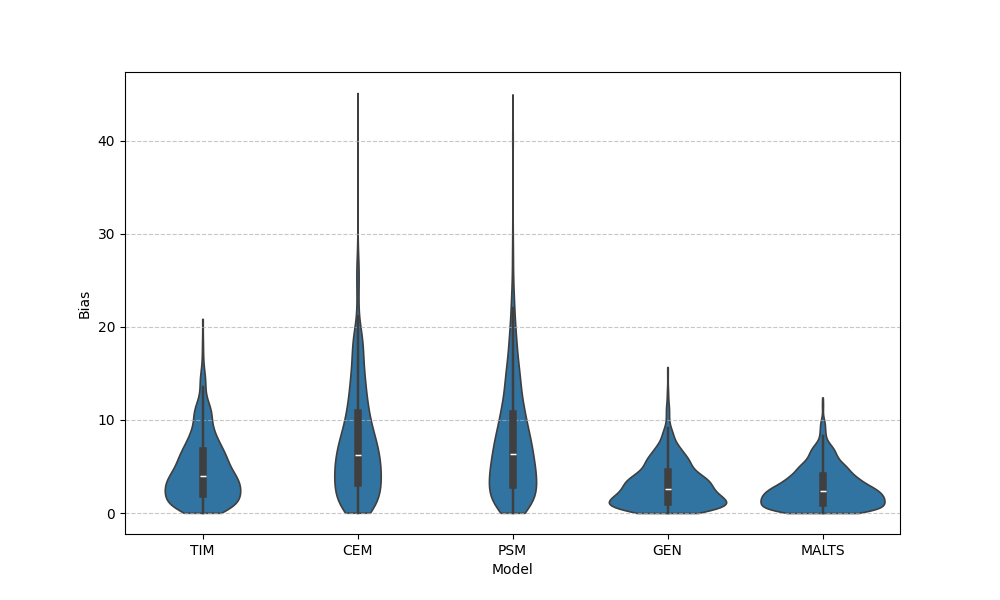}}}\hspace{5pt}
\subfloat[]{%
\resizebox*{8 cm}{!}{\includegraphics{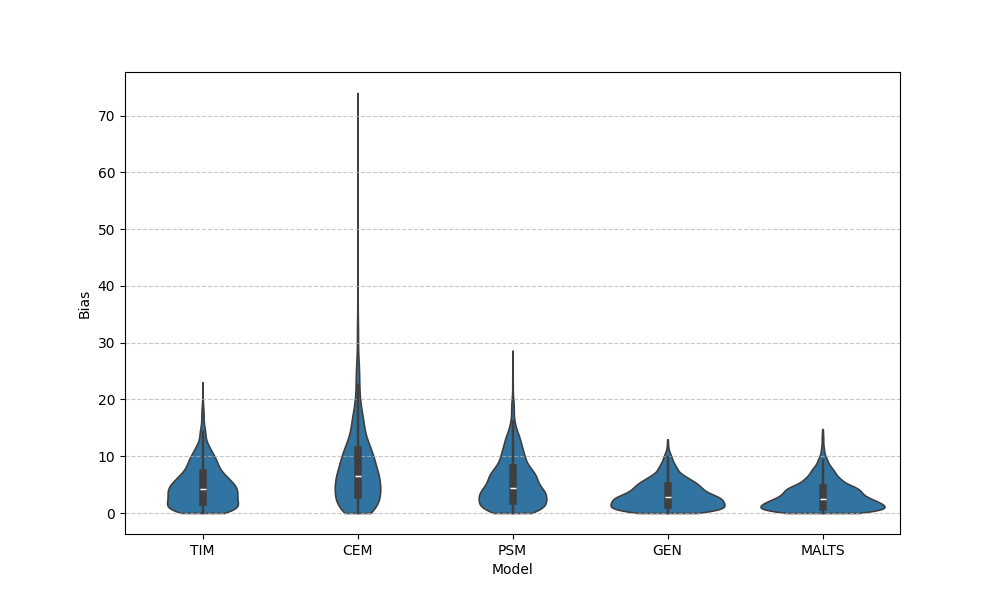}}}\hspace{5pt}
\subfloat[]{%
\resizebox*{8 cm}{!}{\includegraphics{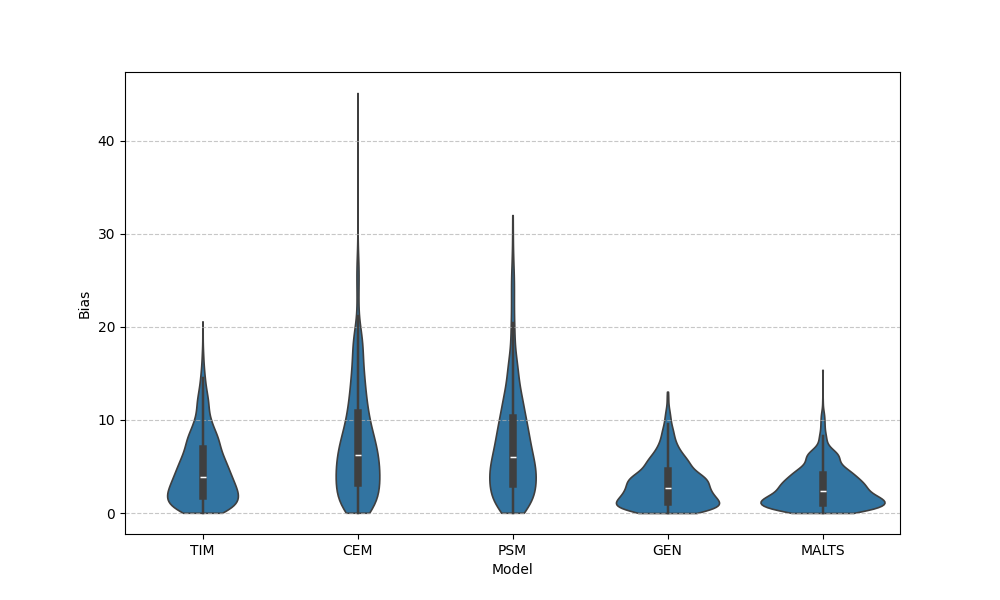}}}\hspace{3pt}

\caption{Bias on CATE calculated scatterplot (a) Scenario 1A (b) Scenario 1B (c) Scenario 2A (d) Scenario 2B (e) Scenario 3A (f) Scenario 3B.} \label{fig:bias}
\end{figure}

The main difference between Scenarios 1, 2, and 3 lies in the association structures, as explained in Section \ref{sec4.1}. To provide a clearer understanding of how the various models perform across these scenarios, we present the bias, \( L_1^m \), and \(  T_{f} \) in Table \ref{biasvar2}, while the time taken by the matching algorithms is shown in Table \ref{time}. Figure \ref{fig:bias} illustrates the distribution of bias across the 1000 datasets run for each scenario.

It is important to note that we also calculated the \( L_1 \) score before matching, which was $1.0$ for all datasets across all scenarios, indicating complete separation in the datasets. From Table \ref{biasvar2}, we observe that CEM achieves perfect overlap with \( L_1^m = 0 \) across all scenarios, 
demonstrating the best performance. TIM, on the other hand, achieves \( L_1^m \) values ranging from 0.372 in 
Scenario 1B to 0.503 in Scenario 2A. This indicates that while CEM ensures complete overlap between treatment 
and control groups—effectively replicating a randomized controlled trial (RCT)—TIM improves balance between 
groups by approximately 50–63\%.  
However, despite CEM’s superior performance in terms of \( L_1^m \), it suffers from a significant limitation: 
the proportion of matched treatment units (\( T_{f} \)) ranges from 0.498 in Scenario 3A to 0.623 in Scenario 1B. 
This implies that, on average, nearly half of the treatment units are unmatched, resulting in considerable data loss. 
In contrast, while TIM does not achieve a perfect \( L_1^m \), it retains all treatment units, ensuring \( T_{f} = 1 \).  
Furthermore, the \( L_1^m \) values for PSM, GEN, and MALTS remain at 1 across all scenarios, indicating no improvement 
in the overlap between the treatment and control distributions. This suggests that PSM, GEN, and MALTS fail to improve balance 
from pre- to post-matching, highlighting TIM’s effectiveness in improving covariate overlap while preserving 
all treatment units.

\renewcommand{\thetable}{\arabic{table}}
\small
\setlength{\tabcolsep}{9pt}
\begin{longtable}{lccccc}

\caption{Bias, $L_1^m$, and $ T_{f}$  of estimated CATE by different variable selection methods.}
\label{biasvar2} \\

\toprule
\textbf{} & \textbf{TIM} & \textbf{CEM} & \textbf{GEN} & \textbf{PSM} & \textbf{MALTS}  \\

\midrule
\endfirsthead
{}{\footnotesize\itshape\tablename~\thetable:
Continued from the previous page} \\
\toprule
\textbf{} & \textbf{TIM} & \textbf{CEM} & \textbf{GEN} & \textbf{PSM} & \textbf{MALTS}  \\
\midrule
\endhead

\midrule
{}{\footnotesize\itshape
} \\
\endfoot
\bottomrule
{\footnotesize\itshape
} \\
\endlastfoot
\textbf{Scenario 1A} & ~ & ~ & ~ & ~ & ~ \\ \hline
        \textbf{$L_1^m$} & 0.499 & 0 & 1 & 1 & 1 \\ 
        \textbf{$ T_{f}$} & 1 & 0.508 & 1 & 1 & 182.698 \\
        \textbf{Bias} & 5.089 & 7.630 & 5.592 & 9.297 & 3.743 \\ 
        \textbf{Lower 95\% CI} & 4.846 & 7.184  & 5.322  & 8.848  & 3.558   \\ 
        \textbf{Upper 95\% CI} & 5.333 &  8.076 & 5.863 &  9.746 & 3.928 \\
        \hline  
            \textbf{Scenario 1B} & ~ & ~ & ~ & ~ & ~  \\ \hline
        \textbf{$L_1^m$} & 0.372 & 0 & 1 & 1 & 1  \\ 
        \textbf{$ T_{f}$} & 1 & 0.623 & 1 & 1 & 169.69 \\
        \textbf{Bias} &  6.838 & 7.618& 3.790 & 10.011 & 3.660 \\ 
        \textbf{Lower 95\% CI} & 6.528 & 7.247  & 3.607  & 9.523  & 3.488    \\ 
        \textbf{Upper 95\% CI} &  7.147 &  7.989 &  3.974 &  10.498&  3.832  \\
        \hline
        \textbf{Scenario 2A} & ~ & ~ & ~ & ~ & ~  \\ \hline
        \textbf{$L_1^m$} & 0.503 & 0 & 1 & 1 & 1  \\ 
        \textbf{$ T_{f}$} & 1 & 0.506 & 1 & 1 & 178.676 \\
        \textbf{Bias} & 4.847  & 8.04 & 3.313 & 5.667 & 3.229  \\ 
       \textbf{Lower 95\% CI} & 4.614 & 7.616 & 3.159 & 5.393 & 3.076  \\
       \textbf{Upper 95\% CI} &  5.079 & 8.470 & 3.467 & 5.941 & 3.381  \\
       \hline
       \textbf{Scenario 2B} & ~ & ~ & ~ & ~ & ~  \\ \hline
        \textbf{$L_1^m$} & 0.390 & 0 & 1 & 1 & 1 \\ 
        \textbf{$ T_{f}$} & 1 & 0.508 & 1 & 1 & 182.698 \\
        \textbf{Bias} & 4.775 & 7.618 & 3.104 & 7.637 & 2.752  \\ 
        \textbf{Lower 95\% CI} & 4.551 & 7.247  & 2.953  & 7.261  & 2.619    \\ 
        \textbf{Upper 95\% CI} &  4.998 &  7.989 &  3.104 &  8.014 &  2.884  \\
        \hline
        \textbf{Scenario 3A} & ~ & ~ & ~ & ~ & ~  \\ \hline
        \textbf{$L_1^m$} & 0.497 & 0 & 1 & 1 & 1\\
        \textbf{$ T_{f}$} & 1 & 0.498 & 1 & 1 & 177.84 \\
        \textbf{Bias} & 4.914 & 8.162 & 3.249 & 5.522 & 3.097\\ 
        \textbf{Lower 95\% CI} & 4.683 & 7.702  & 3.100 & 5.255 & 2.941  \\ 
        \textbf{Upper 95\% CI} &  5.146 & 8.621 & 3.398 & 5.789 & 3.253  \\
        \hline
        \textbf{Scenario 3B} & ~ & ~ & ~ & ~ & ~ \\ \hline
        \textbf{$L_1^m$} & 0.385 & 0 & 1 & 1 & 1  \\ 
        \textbf{$ T_{f}$} & 1 & 0.623 & 1 & 1 & 172.947 \\
        \textbf{Bias} & 4.732 & 7.618 & 3.200 & 7.383 & 2.816   \\ 
       \textbf{Lower 95\% CI} & 4.506 & 7.247  & 3.046 & 7.026  & 2.681   \\
       \textbf{Upper 95\% CI} &  4.958 &  7.989 & 3.353 & 7.739 & 2.952  \\
        
\end{longtable}

Upon further examination of Table \ref{biasvar2}, we find that in terms of bias, MALTS (ranging from 2.752 in Scenario 2B to 3.743 in Scenario 1A) and GEN (ranging from 3.104 in Scenario 2B to 5.592 in Scenario 1A) outperform TIM (ranging from 4.732 in Scenario 3B to 6.832 in Scenario 1B) across all experiments. However, a closer analysis reveals that despite their lower bias, both GEN and MALTS yield \( L_1^m = 1 \), indicating no improvement in overlap between treatment and control groups—a fundamental objective of any matching method \cite{king:2012}.  
In contrast, TIM achieves a better trade-off by outperforming PSM and CEM in terms of bias while also maintaining a significantly lower \( L_1^m \), ensuring improved covariate balance and effective overlap between treatment and control groups.  
Our findings align with the literature \cite{rosenbaum1985constructing}, which suggests that allowing some degree of imprecision in matching while retaining more units leads to better bias than enforcing exact matching on all variables. This is particularly evident in CEM, which, while achieving the best \( L_1^m \) by ensuring perfect overlap, performs poorly in terms of bias.  
  In terms of time, Table \ref{time} shows that while PSM is the fastest, TIM is still more efficient than GEN and MALTS, which are 10 to 30 times slower. Despite its moderate performance in individual areas, TIM successfully balances bias reduction, overlap improvement, and computational efficiency, making it particularly suitable for high-dimensional, mixed-variable healthcare data, where scalability and the ability to improve both bias and data distributions are essential.

\begin{table}[h]
\caption{Average Computational time (seconds) taken by different matching methods to estimate CATE.}\label{time}
\begin{tabular*}{\textwidth}{@{\extracolsep\fill}lccccc}
\toprule%
\textbf{} & \textbf{TIM} & \textbf{CEM} & \textbf{GEN} & \textbf{PSM} & \textbf{MALTS} \\
\midrule
        \textbf{Scenario 1A} & 2.452 & 2.310 & 20.98  & 0.212 &27.83  \\  \hline
        \textbf{Scenario 1B} & 2.335& 2.352 & 44.269 & 0.314 & 33.870  \\ \hline
        \textbf{Scenario 2A} & 3.072 & 1.942 & 42.172 & 0.245 & 29.627  \\  \hline
        \textbf{Scenario 2B} & 3.392 & 2.352 & 55.844 & 0.373 & 41.596  \\ \hline
        \textbf{Scenario 3A} & 2.697 & 2.071 & 41.902  & 0.224 & 61.105 \\  \hline
        \textbf{Scenario 3B} & 2.523 & 2.352 & 60.423 & 0.286 & 33.710 \\ \hline
\end{tabular*}

\end{table}

Our experimental simulation results reveal that while GEN and MALTS achieve the lowest bias in CATE estimation, they perform poorly in terms of computational efficiency and \( L_1^m \). Their high computational cost makes them impractical for large datasets, and their inability to improve \( L_1^m \) from pre-matching to post-matching contradicts the fundamental objective of matching methods. In contrast, TIM consistently outperforms CEM and PSM—two widely regarded matching techniques—in estimating CATE while also meeting theoretical expectations by improving \( L_1^m \) in the post-matching sample. While other methods exhibit trade-offs in either bias, balance, or scalability, TIM provides a well-rounded approach that effectively balances estimation quality and computational efficiency. 
 Although TIM may not be the best across all individual measures, it can certainly be considered the "jack of all trades" in matching methods.

\section{Case Study: Does high cholesterol lead to diabetes?} \label{sec5}

In this section, we evaluate the causal relationship between high cholesterol and diabetes and demonstrate the efficiency of the proposed matching framework in improving $L_1^m$ and calculating the Conditional Average Treatment Effect (CATE) for large-scale data. For this case study, we use the CDC Diabetes Health Indicators dataset, which was made available on Kaggle in 2023. This dataset is a cleaned version of the Behavioral Risk Factor Surveillance System (BRFSS) \citep{BRFSS}, a health-related telephone survey conducted annually by the CDC. Each year, the BRFSS collects responses from over 400,000 Americans regarding health-related risk behaviors, chronic health conditions, and the use of preventive services. The dataset has been cleaned according to the methodology outlined in \cite{xie2019building}.

\subsection{Data Description and Results}

\begin{sidewaystable}
\small
\caption{Socioeconomic and health characteristics of participants in CDC Data. The number in parentheses represents the percentage of the sample.}\label{CDC}
\begin{adjustbox}{scale=0.8,center}
\begin{tabular*}{\textheight}{@{\extracolsep\fill}llllll}
\toprule%
        \textbf{Covariates} & \textbf{All Participants} & \textbf{No Diabetes} & \textbf{Diabetes} & \textbf{No High Cholesterol}  & \textbf{No High Cholesterol} \\
\midrule
 \textbf{Age} & ~ & ~ & ~ & ~ & ~ \\ \hline
        18-25 &  5700 (2.24) &	5622 (98.63)&	78 (1.36)&	5198 (91.19)&	502 (8.80)
\\ 
        26-34 & 18721 (7.37)	&18267 (97.57) &	454 (2.42) 	&15940 ( 85.14)&	2781 (14.85) \\ 
        35-49 & 49799 (19.63) &	46380	(93.13)&3419(6.86)&	36002(72.29)&	13797(27.70)
 \\ 
        50+ & 179460 (70.74) &	148065 (82.50)&	31395 (17.49) &	88949 (49.56) &	90511 (50.43)
 \\ \hline
        \textbf{Gender}  & ~ & ~ & ~ & ~ & ~ \\ \hline
        Male  & 111706 (44.03) &	94771 (87.03)&	16935 (12.96)&	62387 (58.95)&	49319 (41.04)
 \\ 
        Female  & 141974 (55.96) &	123563 (84.83)&	18411 (15.16)&	83702 (55.84)&	58272 (44.15)
 \\ \hline
        \textbf{Education}  & ~ & ~ & ~ & ~ & ~ \\ \hline
        Never Attended School  & 174 (0.06)&	127 (72.98)&	47 (27.01)&	87 (0.50)&	87 (0.50)
 \\ 
        Grades 1 through 8  & 4043 (1.59) &	2860 (70.73)&	1183 (29.26)&	1878 (46.45)&	2165 (53.54)
 \\ 
        Grades 9 through 11 & 9478 (3.73)&	7182 (75.77)&	2296 (24.22)&	4741 (50.02)&	4737 (49.97)
 \\ 
        Grade 12 or GED & 62750 (24.73) &	51684 (82.36) &	11066 (17.6)&	33789 (53.84) &	28961 (46.15)
 \\ 
         College 1 to 3 years  & 69910 (27.55) &	59556 (85.18) &	10354 (14.81) &	40223 (57.53) &	29687 (42.46)
 \\ 
        College graduate  & 107325 (42.30) &	96925 (90.30) &	10400 (9.69)&	65371 (60.90) &	41954 (39.09) 
 \\ 
\hline
        \textbf{Income} & ~ & ~ & ~ & ~ & ~ \\ \hline
        Less than \$25,000 &94122 (37.10)& 37588 (39.93)&	28551 (30.33)&	9037 (9.60)&	18946(20.12)
 \\ 
        \$25,000 - \$34,999 & 116968 (46.10) & 46018 (39.34)&	37460 (32.02)&	8558 (7.31) &	24932 (21.31)
 \\ 
        \$35,000 - \$54,999 & 36470 (14.37)&	31179 (85.49)&	5291 (14.50)&	20610 (56.51) &	15860 (43.48)
 \\ 
        \$55,000 - \$74,999 & 43219 (17.03)&	37954 (87.81) &	5265 (12.18)&	25007 (57.86)&	18212 (42.13)
 \\ 
        \$75,000 or more  & 90385 (35.62)&	83190 (92.03)&	7195 (7.96)&	56594 (62.61)&	33791 (37.38)
 \\ \hline
        \textbf{Heavy Alcohol Consumption}   & ~ & ~ & ~ & ~ & ~ \\ \hline
        Yes & 14256 (5.61) &13424 (94.16) &	832 (5.83) &	8543 (59.92) &	5713 (40.07) 
 \\ \
        No  & 239424 (94.38) &	204910 (85.58) &	34514 (14.41) &	137546 (57.44)&	101878 (42.55)
 \\ 
\hline
        \textbf{Heart Disease or Attack}  & ~ & ~ & ~ & ~ & ~ \\ \hline
        Yes  & 23893 (9.41)&	16015 (67.02)&	7878 (32.97)&	7140 (29.88)&	16753 (70.11)
 \\ 
        No & 229787 (90.58) &	202319 (88.04) &	27468 (11.95)&	138949 (60.46)&	90838 (39.53)
 \\ 
 \\ \hline
        \textbf{Physical Activity in past 30 days}  & ~ & ~ & ~ & ~ & ~ \\ \hline
        Yes  & 191920 (75.65)&	169633 (88.38)&	22287 (11.61)&	114722 (59.77) &	77198 (40.22)
 \\ 
        No & 61760 (24.34) &	48701 (78.85) &	13059 (21.14) & 	31367 (50.78) &	30393 (49.21)
 \\ 
 \\ \hline
\end{tabular*}
\end{adjustbox}
\end{sidewaystable}

\begin{table}[h]
\caption{Results for the CDC case study.}\label{CDC_results}
\begin{tabular*}{\textwidth}{@{\extracolsep\fill}lccccc}
\toprule%
\textbf{} & \textbf{TIM} & \textbf{CEM} & \textbf{GEN} & \textbf{PSM} & \textbf{MALTS} \\
\midrule
        \textbf{CATE} & 0.079 & 0.034 & -  & 0.009 &-  \\  \hline
        \textbf{$L_1$} & 0.916 & 0.916 & -  & 0.916 & -  \\  \hline
        \textbf{$L_1^m$} & 0.654& 0 & -  & 0.932 & -  \\ \hline
        \textbf{$T_f$} & 1 & 0.34 & -  & 1 & -  \\ \hline
        \textbf{Time} & 362 min & 48 min  &  > 3 days  & 37 min &-  \\  \hline

\end{tabular*}

\end{table}

Our dataset comprises 253,680 samples from 2015, including 22 covariates. Among these samples, 35,346 individuals were diagnosed with diabetes. The dataset contains information on socioeconomic factors, health status, education, income, and mental health. It includes mixed data types, with continuous variables such as BMI and ranges for various health indicators, including mental, physical, and general health. Additionally, it features discrete variables for factors such as smoking, heart disease, and heavy alcohol consumption.  

A summary of the socioeconomic and health characteristics of the participants is presented in Table \ref{CDC}. We apply all previously discussed matching methods from the numerical experiments section. Since this is observational data, we cannot directly compute bias, as the true treatment effect is unknown. However, following the theoretical expectations of matching, we compute \( L_1 \) and \( L_1^m \) to assess the improvement in overlap from pre-matching to post-matching, and $T_f$.

From Table \ref{CDC_results}, similar to the numerical experiments, we observe that CEM with $T_f = 0.34$ fails to include most of the treatment groups for matching. PSM worsens the balance after matching ($L_1^m > L_1$), while TIM improves the balance post-matching and successfully matches all treatment units with $T_f = 1$. Due to the high-dimensional nature of the data, MALTS fails to converge, and GEN does not complete even after running the algorithm for over 3 days. Additionally, TIM estimates a Conditional Average Treatment Effect (CATE) of 0.079 for the effect of high cholesterol on diabetes, which is consistent with the estimates from other methods such as CEM and PSM. This positive causal effect indicates that individuals with high cholesterol are more likely to develop diabetes, although the small effect size limits its clinical significance.

\section{Discussion} \label{sec6}

To evaluate TIM, we used synthetic data based on three scenarios outlined in Section \ref{sec4}. Each scenario represented varying strengths of confounders, resembling different situations encountered in real-world datasets, with correlations of $\rho = 0$ and $\rho = 0.5$ representing no correlation and high correlation, respectively. In the synthetic dataset, the true CATE and initial overlap ($L_1$) were known. The goal was to minimize bias while improving the overall multivariate balance pre- and post-matching ($L_1^m < L_1$), which the synthetic data allowed us to assess. Our findings indicate that TIM outperforms CEM and PSM in terms of bias but performs worse than GEN and MALTS. However, TIM surpasses GEN and MALTS in improving overlap and computational efficiency. This advantage holds across all scenarios, demonstrating that TIM is well-suited for high-dimensional datasets, such as those encountered in healthcare settings.

We also validated the applicability of TIM using a real-world dataset from the CDC. The estimated Conditional Average Treatment Effect (CATE) was 0.079, suggesting that high cholesterol contributes to a 7.9\% increase in the likelihood of developing diabetes. This result aligns with the medical literature, including research by \cite{tajima2014high}. 
The dataset used in this study has several limitations. The BRFSS is a self-reported survey, which introduces the risk of recall bias and under- or over-reporting. Additionally, since the data is cross-sectional, establishing temporal relationships—crucial for making robust causal conclusions—remains challenging. Unobserved confounders, which are not fully accounted for in the dataset, further limit the accuracy of the estimates. Given these limitations, we caution against drawing definitive clinical or policy-related conclusions from this study. 
The authors aim to develop and refine causal inference techniques to support healthcare practitioners. Collaborating with medical experts can help fine-tune these models, ensuring that all confounders are addressed and results are reliable. To create actionable healthcare policies, it is crucial to use multiple high-quality datasets and collaborate with medical experts to draw robust and reliable conclusions.

For future research, we plan to incorporate multiple treatment levels to address more complex treatment regimes. Additionally, we aim to optimize the algorithm to reduce computation time for high-dimensional datasets, with a focus on enhancing computational efficiency. We also intend to explore case studies in other domains, such as economics and social sciences, where policy evaluation plays a critical role. This will allow us to test the generalizability of our approach across various fields.

\section{Conclusion} \label{sec7}
In this paper, we introduce a novel matching algorithm aimed at improving the balance from pre- to post-matching. While existing matching methods focus on reducing bias, none specifically prioritize enhancing the overall multivariate balance between control and treatment groups. To address this gap, we propose the Two-Stage Interpretable Matching Framework (TIM), which first matches on all variables and then iteratively reduces the number of variables based on confounder importance. In the second stage, we refine the matches by learning a mixed distance metric to calculate the Conditional Average Treatment Effect (CATE). We evaluate TIM based on bias, multivariate overlap, and computational time. Our results show that TIM outperforms commonly used methods, such as Coarsened Exact Matching (CEM) and Propensity Score Matching (PSM), in at least one aspect of performance. While TIM improves balance within the dataset, it is currently limited to binary treatment and cannot be applied to longitudinal data. Finally, we demonstrate the application of our method to real-world data. Our analysis suggests that individuals with high cholesterol are more susceptible to developing diabetes, with performance in the real-world setting aligning closely with the results from the simulation experiments. However, due to the limitations of the dataset, the estimated causal effect should be interpreted with caution.

\section*{Data Availability Statement}
This study uses two sets of experiments to support the findings.  The first set consists of simulated data, and section \ref{sec4} describes how to replicate the experiments. The second set of experiments uses the Behavioral Risk Factor Surveillance System (BRFSS) data from the Centers for Disease Control and Prevention (CDC). This dataset is publicly available from CDC website at "https://www.cdc.gov/brfss/annual\_data/annual\_2015.html"\citep{BRFSS}.

\section*{Disclosure statement}

The authors report there are no competing interests to declare.

\section*{Funding}

Financial support from the National Science Foundation (Award Number:
2047094) is greatly acknowledged.

\section*{Consent and Approval Statement}

Not Applicable, since our study utilizes publicly available data. 

\bibliographystyle{apalike} 
\bibliography{bib}

\appendix

\section{Additional Numerical Experiments} \label{s:intro}

We replicate the data-generating equations from the manuscript and conduct an experiment with a larger dataset of \( n = 4000 \) and 16 covariates, consisting of 10 continuous and 6 categorical variables. As before, we consider three scenarios with differing association structures. Each scenario is further divided into two subsections (A and B) based on correlation levels (\(\rho = 0\) and \(\rho = 0.5\)). The results of these experiments are presented in Table \ref{appendixbiasvar2} and \ref{appentime}.

\textbf{Scenario 4:} Includes strong confounders.
\begin{itemize}
    \item Treatment Assignment: $T \sim Bernoulli( p = Expit(0.8X_{c_1} + \cdots + 0.8X_{c_{10}} +0.8X_{d_1}+ \cdots + 0.8X_{d_6}))$ where, $Expit (x) = \frac{e^x}{1+e^x} $
    \item Outcome Model: The outcome variable $Y$ is generated as
    $y = TE.T + 0.8X_{c_1} + \cdots + 0.8X_{c_{10}} +0.8X_{d_1}+ \cdots + 0.8X_{d_6} + \epsilon$,
    where $\epsilon \sim \mathcal{N}(0, \sigma_x)$ represents random noise and $TE = 1$ is the treatment effect and $T=0$ when unit does not receive treatment, $T=1$ when it receives treatment.
    \item Correlation:  $\rho = 0$ (Scenario 4A),$\rho = 0.5$ (Scenario 5B) 
\end{itemize}

\textbf{Scenario 5:}  Includes confounders with varying strengths—strong, medium, and weak.
\begin{itemize}
    \item Treatment Assignment: $T \sim Bernoulli( p = Expit(0.8X_{c_1} +\cdots+ 0.8X_{c_4}+0.5X_{c_5}+\cdots+0.5X_{c_8} + 0.2X_{c_9} + 0.2X_{c_{10}} +0.8X_{d_1}+ 0.8X_{d_2}+ 0.5X_{d_3} + 0.5X_{d_4} + 0.2X_{d_5}+0.2X_{d_6})$ where, $Expit (x) = \frac{e^x}{1+e^x} $
    \item Outcome Model: The outcome variable $Y$ is generated as
    $y = TE.T + 0.8X_{c_1} +\cdots+ 0.8X_{c_4}+0.5X_{c_5}+\cdots+0.5X_{c_8} + 0.2X_{c_9} + 0.2X_{c_{10}} +0.8X_{d_1}+ 0.8X_{d_2}+ 0.5X_{d_3} + 0.5X_{d_4} + 0.2X_{d_5}+0.2X_{d_6} + \epsilon$,
    where $\epsilon \sim \mathcal{N}(0, \sigma_x)$ represents random noise and $TE = 1$ is the treatment effect and $T=0$ when unit does not receive treatment, $T=1$ when it receives treatment.
    \item Correlation:  $\rho = 0$ (Scenario 5A),$\rho = 0.5$ (Scenario 5B) 
\end{itemize}

\textbf{Scenario 6:}  Includes strong confounders, medium confounders, and confounders that are strongly associated with the treatment.
\begin{itemize}
    \item Treatment Assignment: $T \sim Bernoulli( p = Expit(0.8X_{c_1} +\cdots+ 0.8X_{c_4}+0.5X_{c_5}+\cdots+0.5X_{c_8} + 0.8X_{c_9} + 0.8X_{c_{10}} +0.8X_{d_1}+ 0.8X_{d_2}+ 0.5X_{d_3} + 0.5X_{d_4} + 0.8X_{d_5}+0.8X_{d_6})$ where, $Expit (x) = \frac{e^x}{1+e^x} $
    \item Outcome Model: The outcome variable $Y$ is generated as
    $y = TE.T +0.8X_{c_1} +\cdots+ 0.8X_{c_4}+0.5X_{c_5}+\cdots+0.5X_{c_8} + 0.2X_{c_9} + 0.2X_{c_{10}} +0.8X_{d_1}+ 0.8X_{d_2}+ 0.5X_{d_3} + 0.5X_{d_4} + 0.2X_{d_5}+0.2X_{d_6} + \epsilon$,
    where $\epsilon \sim \mathcal{N}(0, \sigma_x)$ represents random noise and $TE = 1$ is the treatment effect and $T=0$ when unit does not receive treatment, $T=1$ when it receives treatment.
    \item Correlation:  $\rho = 0$ (Scenario 6A),$\rho = 0.5$ (Scenario 6B) 
\end{itemize}

\renewcommand{\thetable}{A\arabic{table}}
\small
\setlength{\tabcolsep}{9pt}
\begin{longtable}{lccccc}

\caption{Bias, $L_1^m$, and $ T_{f}$  of estimated CATE by different variable selection methods.}
\label{appendixbiasvar2} \\

\toprule
\textbf{} & \textbf{TIM} & \textbf{CEM} & \textbf{GEN} & \textbf{PSM} & \textbf{MALTS}  \\

\midrule
\endfirsthead
{}{\footnotesize\itshape\tablename~\thetable:
Continued from the previous page} \\
\toprule
\textbf{} & \textbf{TIM} & \textbf{CEM} & \textbf{GEN} & \textbf{PSM} & \textbf{MALTS}  \\
\midrule
\endhead

\midrule
{}{\footnotesize\itshape
} \\
\endfoot
\bottomrule
{\footnotesize\itshape
} \\
\endlastfoot
\textbf{Scenario 4A} & ~ & ~ & ~ & ~ & ~ \\ \hline
        \textbf{$L_1^m$} & 0.987 & 0 & 1 & 1 & 1 \\ 
        \textbf{$ T_{f}$} & 1 & 0.613 & 1 & 1 & 175.740 \\
        \textbf{Bias} & 7.310 & 9.338 & 7.462 & 9.215 & 3.769 \\ 
        \textbf{Lower 95\% CI} & 3.775 & 4.528  & 3.512  & 5.974  &2.528   \\ 
        \textbf{Upper 95\% CI} & 10.845 &  12.195 & 11.413 &  12.455 & 5.009 \\
        \hline  
            \textbf{Scenario 4B} & ~ & ~ & ~ & ~ & ~  \\ \hline
        \textbf{$L_1^m$} & 0.948 & 0 & 1 & 1 & 1  \\ 
        \textbf{$ T_{f}$} & 1 & 0.077 & 1 & 1 & 175.715 \\
        \textbf{Bias} &  12.875 & 10.916& 6.990 & 15.446 & 6.223 \\ 
        \textbf{Lower 95\% CI} & 6.812 & 4.694& 2.809  & 11.886  & 4.009    \\ 
        \textbf{Upper 95\% CI} &  18.938 &  17.138 & 11.172 & 19.007&  8.458  \\
        \hline
        \textbf{Scenario 5A} & ~ & ~ & ~ & ~ & ~  \\ \hline
        \textbf{$L_1^m$} & 0.987 & 0 & 1 & 1 & 1  \\ 
        \textbf{$ T_{f}$} & 1 & 0.012 & 1 & 1 & 177.529 \\
        \textbf{Bias} & 3.777  & 34.770 & 7.222 & 6.676 & 1.942  \\ 
       \textbf{Lower 95\% CI} & 2.636 & 23.528 & 4.545 & 2.969& 1.384  \\
       \textbf{Upper 95\% CI} &  4.917 & 46.013 & 9.898 &10.383 & 2.501  \\
       \hline
       \textbf{Scenario 5B} & ~ & ~ & ~ & ~ & ~  \\ \hline
        \textbf{$L_1^m$} & 0.922 & 0 & 1 & 1 & 1 \\ 
        \textbf{$ T_{f}$} & 1 & 0.049 & 1 & 1 & 179.425 \\
        \textbf{Bias} & 4.564 & 24.022 & 6.990 & 16.583 & 5.016 \\ 
        \textbf{Lower 95\% CI} & 3.068 & 6.842  & 2.809  & 10.411  & 2.682   \\ 
        \textbf{Upper 95\% CI} &  6.061 &  41.202 &  11.172 &  22.755 &  7.349  \\
        \hline
        \textbf{Scenario 6A} & ~ & ~ & ~ & ~ & ~  \\ \hline
        \textbf{$L_1^m$} & 0.989 & 0 & 1 & 1 & 1\\
        \textbf{$ T_{f}$} & 1 & 0.010 & 1 & 1 & 185.974 \\
        \textbf{Bias} & 4.395 & 28.649 & 6.251 & 9.290 & 2.994\\ 
        \textbf{Lower 95\% CI} & 2.467 & 11.383  & 2.796 & 2.976 & 1.958  \\ 
        \textbf{Upper 95\% CI} &  6.323 & 45.914 & 9.706 & 15.603 & 4.030  \\
        \hline
        \textbf{Scenario 6B} & ~ & ~ & ~ & ~ & ~ \\ \hline
        \textbf{$L_1^m$} & 0.962 & 0 & 1 & 1 & 1  \\ 
        \textbf{$ T_{f}$} & 1 & 0.051 & 1 & 1 & 176.703 \\
        \textbf{Bias} & 6.525 & 20.979 & 4.546 & 15.117 & 5.676   \\ 
       \textbf{Lower 95\% CI} & 4.294 & 9.089  & 1.906 & 8.845  & 4.430   \\
       \textbf{Upper 95\% CI} &  8.756 &  32.868 & 7.186 & 21.388 & 6.921  \\
        
\end{longtable}

\begin{table}[h]
\caption{Average Computational time (seconds) taken by different matching methods to estimate CATE.}\label{appentime}
\begin{tabular*}{\textwidth}{@{\extracolsep\fill}lccccc}
\toprule%
\textbf{} & \textbf{TIM} & \textbf{CEM} & \textbf{GEN} & \textbf{PSM} & \textbf{MALTS} \\
\midrule
        \textbf{Scenario 4A} & 18.169 & 2.075 & 82.49  & 0.089 &249.332 \\  \hline
        \textbf{Scenario 4B} & 15.503& 7.711 & 73.402 & 0.104 & 205.880 \\ \hline
        \textbf{Scenario 5A} & 18.867 & 7.366 & 68.304 & 0.105 & 260.165  \\  \hline
        \textbf{Scenario 5B} & 15.745 & 7.587 & 73.402 & 0.106 & 284.983  \\ \hline
        \textbf{Scenario 6A} & 21.755 & 7.847 & 48.387  & 0.082 & 264.890 \\  \hline
        \textbf{Scenario 6B} & 16.197 & 7.659 & 52.218 & 0.084 & 243.236 \\ \hline
\end{tabular*}

\end{table}

The results from Scenarios 4–6 align with the findings in the main manuscript, confirming TIM’s ability to balance bias reduction, computational efficiency, and multivariate overlap. Across all scenarios, TIM consistently achieved lower bias than CEM and PSM while retaining a higher proportion of treatment units. Moreover, TIM improved post-matching overlap, as indicated by its high $L_1^m$ scores (e.g., 0.987 in Scenario 5A), whereas CEM achieved perfect balance at the expense of severe sample exclusion. In terms of computational efficiency, TIM significantly outperformed MALTS, proving to be more than ten times faster on average.

\end{document}